\pdfoutput=1

\documentclass[11pt]{article}

\usepackage[preprint]{acl}

\usepackage{times}
\usepackage{latexsym}
\usepackage{pifont}
\usepackage[T1]{fontenc}

\usepackage[utf8]{inputenc}
\usepackage{amssymb}
\usepackage{microtype}
\usepackage{listings}
\usepackage{xcolor}

\definecolor{codegreen}{rgb}{0,0.6,0}
\definecolor{codegray}{rgb}{0.5,0.5,0.5}
\definecolor{codepurple}{rgb}{0.58,0,0.82}
\definecolor{backcolour}{rgb}{0.95,0.95,0.92}

\lstdefinestyle{mystyle}{
    backgroundcolor=\color{backcolour},   
    commentstyle=\color{codegreen},
    keywordstyle=\color{magenta},
    numberstyle=\tiny\color{codegray},
    stringstyle=\color{codepurple},
    basicstyle=\ttfamily\footnotesize,
    breakatwhitespace=false,         
    breaklines=true,                 
    captionpos=b,                    
    keepspaces=true,                 
    numbers=left,                    
    numbersep=5pt,                  
    showspaces=false,                
    showstringspaces=false,
    showtabs=false,                  
    tabsize=2
}

\usepackage{inconsolata}

\usepackage{graphicx}

\title{MC-MKE: A Fine-Grained Multimodal Knowledge Editing Benchmark Emphasizing Modality Consistency}

\author{Junzhe Zhang, Huixuan Zhang, Xunjian Yin, Baizhou Huang, Xu Zhang, Xinyu Hu, Xiaojun Wan\\ 
Wangxuan Institute of Computer Technology, Peking University \\
         \{junzhezhang, zhanghuixuan\}@stu.pku.edu.cn  \\ \{xjyin, hbz19, zhangxu, huxinyu, wanxiaojun\}@pku.edu.cn}

\usepackage{color}

\usepackage{amsmath}
\usepackage{bbm}
\usepackage{multirow}
\usepackage{graphicx}
\usepackage{array}
\usepackage{booktabs}
\usepackage{bbding}
\usepackage{amssymb}
\usepackage{makecell}
\usepackage{enumitem}
\graphicspath{{figures/}}

\begin{document}
\maketitle
\begin{abstract}
Multimodal large language models (MLLMs) are prone to non-factual or outdated knowledge issues, which can manifest as misreading and misrecognition errors due to the complexity of multimodal knowledge. Previous benchmarks have not systematically analyzed the performance of editing methods in correcting these two error types. To better represent and correct these errors, we decompose multimodal knowledge into its visual and textual components. Different error types correspond to different editing formats, which edit distinct parts of the multimodal knowledge. We present MC-MKE, a fine-grained \underline{\textbf{M}}ultimodal \underline{\textbf{K}}nowledge \underline{\textbf{E}}diting benchmark emphasizing \underline{\textbf{M}}odality \underline{\textbf{C}}onsistency. Our benchmark facilitates independent correction of misreading and misrecognition errors by editing the corresponding knowledge component. We evaluate four multimodal knowledge editing methods on MC-MKE, revealing their limitations, particularly in terms of modality consistency. Our work highlights the challenges posed by multimodal knowledge editing and motivates further research in developing effective techniques for this task.
\end{abstract}

\section{Introduction}







\begin{figure}
    \centering
    \includegraphics[width=8cm]{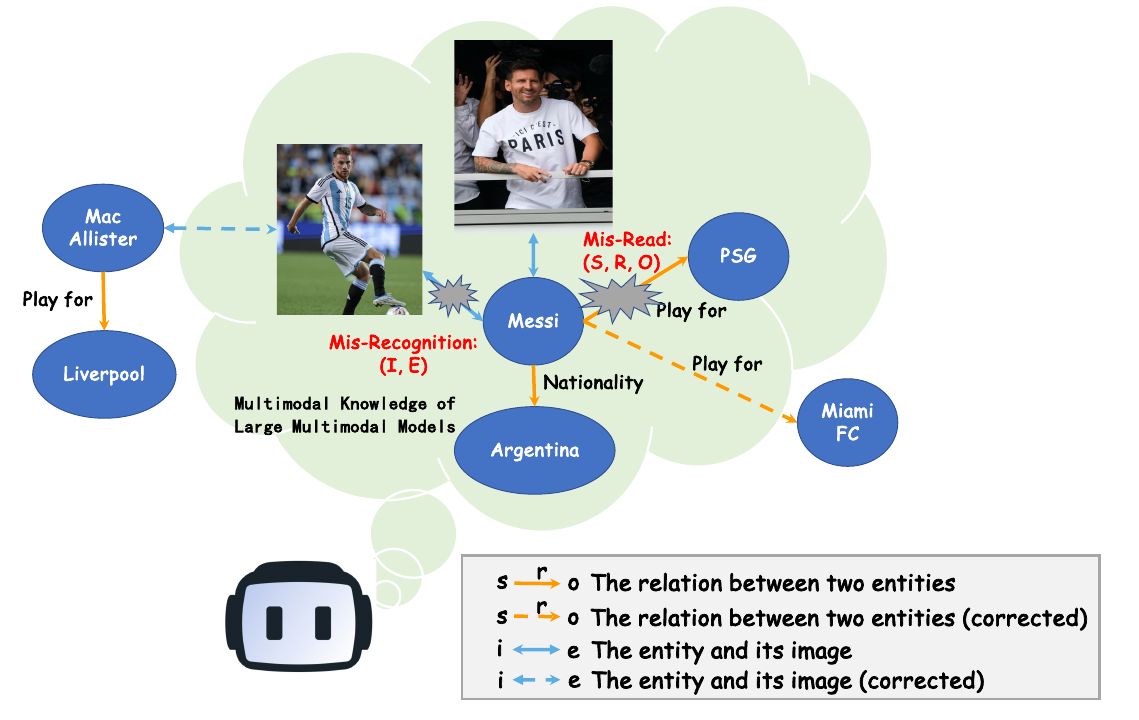}
    \caption{An illustration of multimodal knowledge and the two types of multimodal errors: misrecognizing a picture of Mac Allister as Messi, and misreading Messi's football team.}
    \label{fig:intro}
    \vspace{-15pt}
\end{figure}

With the developments of multimodal large language models (MLLMs), their application has become widespread across various fields.
However, these models struggle with the challenge that the knowledge stored within them could be inaccurate or outdated.
This issue manifests in two errors: misreading and misrecognition \cite{cheng2024edit}. As shown in Figure \ref{fig:intro}, misrecognition occurs when a model mistakenly identifies an image, such as mistaking Mac Allister as Messi. On the other hand, misreading refers to incorrect textual knowledge, such as misremembering Messi's football team. Recent researches\cite{cheng2024edit} have introduced knowledge editing in multimodal contexts to address these issues. 


Following the conventional definition of knowledge-editing in LLMs, a few studies have proposed benchmarks for knowledge editing in MLLMs \cite{cheng2024edit,huang2024kebench,li2024mike}. However, these benchmarks over-simplify the evaluation of multimodal knowledge editing, and do not distinguish the differences between misreading and misrecognition errors\cite{cheng2024edit, huang2024kebench}.
Mixing evaluation of the two types of errors leads to inaccurate assessments of knowledge editing methods in real-world scenarios. Methods may appear to inject objective multimodal knowledge successfully but actually conduct incorrect edits. Take the misreading error in Figure \ref{fig:intro} for an example, where an MLLM misrecognizes the image of Messi to Mac Allister, leading to the erroneous knowledge that "the person in the image plays for Liverpool". If a knowledge editing method falsely injects a knowledge triplet (\textit{Mac Allister, Play for, Inter Miami}), it may still achieve great performance on prior benchmarks, since the multimodal knowledge (\textit{Image of Messi, Play for, Inter Miami}) is actually corrected.

To better handle and evaluate these two types of knowledge editing scenarios, we for the first time define multimodal knowledge in a decomposed format consisting of visual knowledge and textual knowledge in multimodal knowledge editing task. 
In this way, the misreading and misrecognition errors can be distinguished, and thereby be independently corrected by editing different knowledge components. 
The decomposition of multimodal knowledge also brings up another requirement  \textbf{Consistency}. We believe that a knowledge editing method should always ensure the consistency of knowledge across different modalities.
This property is the essential difference between multimodal knowledge editing and uni-modal knowledge editing.




Following the decomposed definition of multimodal knowledge, we propose a multimodal knowledge editing benchmark emphasizing modality consistency (\textbf{MC-MKE}). MC-MKE consists of three subsets, corresponding to the three different formats of multimodal knowledge.
Our benchmark aligns more closely with multimodal knowledge editing in real-life scenarios and can more systematically and comprehensively evaluate the performance of a multimodal knowledge editing method in a fine-grained manner.

We evaluate four of the most renowned multimodal knowledge editing methods including fine-tuning, MEND~\cite{mitchell2022fast}, IKE~\cite{zheng2023edit}, and SERAC~\cite{pmlr-v162-mitchell22a} on the three subsets of different editing formats. We find that the performance of these methods is far from satisfaction on MC-MKE. None of them can achieve great performance on all three different editing formats, especially for the consistency metric. 
It is demonstrated that multimodal knowledge editing is still challenging and requires further exploration.


%




In summary, our contributions are as follows\footnote{Our code and data will be released to the community to facilitate future research.}:
\begin{itemize}[leftmargin=*,itemsep=1pt]
    \item We first propose a decomposed definition of multimodal knowledge according to different multimodal knowledge error types in multimodal knowledge editing task. 
    \item We present \textbf{MC-MKE}, a new multimodal knowledge editing benchmark that can evaluate Reliability, Locality, Generality, and Consistency of multimodal editing methods under different editing formats.
    \item We conduct experiments with various knowledge editing methods on \textbf{MC-MKE}. The results reveal the limitations of existing methods, especially for modality consistency. \textbf{Different} from previous research, we find that editing the corresponding component sometimes yields better performance.
\end{itemize}

\begin{table*}
\centering
\small
\setlength{\tabcolsep}{4pt}
\renewcommand{\arraystretch}{1.25}
\begin{tabular}{cccccccccc}
\toprule
\multirow{2}*{\textbf{Benchmark}}& \multicolumn{3}{c}{\textbf{Edit\_formats}} &  & \multicolumn{5}{c}{\textbf{Edit\_requirements}}  \\ \cline{2-4} \cline{6-10}
 & \textbf{IE} & \textbf{SRO} & \textbf{IRO} & \textbf{Fine-grained} & \textbf{Reliability} & \textbf{Locality} & \textbf{Generality} &  \textbf{Portability} & \textbf{Consistency} \\
\midrule
MMEdit  & \ding{55} & \ding{55} & \ding{51} & \ding{55} & \ding{51} & \ding{51} & \ding{51} & \ding{55} & \ding{55} \\
KEBench & \ding{51} & \ding{55} & \ding{55} & \ding{51} & \ding{51} & \ding{51} & \ding{51} & \ding{51} &  \ding{55} \\
MIKE & \ding{51} & \ding{55} & \ding{55} & \ding{51} & \ding{51} &  \ding{51} & \ding{51} & \ding{55} & \ding{55} \\
MC-MKE & \ding{51} & \ding{51} & \ding{51} & \ding{51} & \ding{51} & \ding{51} & \ding{51} & \ding{51} & \ding{51} \\
\bottomrule
\end{tabular}
\caption{Comparisons of current multimodal knowledge editing benchmarks, MMEdit~\citep{cheng2024edit}, KEBench~\citep{wu2024updating} and MIKE~\citep{li2024mike}. IE, SRO, and IRO represent different editing formats. \ding{51} and \ding{55} mean whether the benchmark can provide data of corresponding editing format. In Fine-grained, \ding{51} means that the corresponding benchmark is constructed based on fine-grained entity information, while \ding{55} means that the benchmark is constructed around multimodal task data. Edit\_requirements are the properties we expect from a good editing method. \ding{51} and \ding{55} indicate whether the benchmark contains the ability to test these properties of editing methods.}
\label{tab:1}
\vspace{-10pt}
\end{table*}

\section{Related Works}
\subsection{Knowledge Editing}
Knowledge editing aims to provide efficient and lightweight solutions for updating knowledge in models \citep{zhu2020modifying}. Several benchmarks have been developed for this task, including COUNTERFACT \citep{meng2022locating} for counterfactual knowledge, MQuake \citep{zhong2023mquake} for multi-hop knowledge, AToKE \citep{yin2024history} for retaining old knowledge, and WIKIUPDATE \citep{wu2024updating} for unstructured knowledge.

These benchmarks primarily address language model editing, leaving multimodal model editing underexplored. To address this gap, \citet{cheng2024edit} introduced the MMEdit benchmark based on Visual QA \citep{7410636} and Image Captioning \citep{NEURIPS2019_680390c5}. \citet{wu2024updating} developed KEBench, which uses multimodal Knowledge Graphs \citep{liu2019mmkg} to evaluate vision knowledge editing. Additionally, MIKE \citep{li2024mike} focuses on fine-grained multimodal entity knowledge editing.
However, as shown in Table~\ref{tab:1}, all previous work has neglected the organization of multimodal knowledge and lacked a more careful definition of multimodal knowledge editing, which is what our work focuses on.



\subsection{Multimodal Models}
Multimodal large language models have developed rapidly in recent years. BLIP-2 \citep{li2023blip} apply Q-Former architecture to transform image input into LLMs input tokens. LLaVA\citep{liu2024visual} and LLaVA-v1.5\citep{liu2024improved} utilize linear layers or perceptrons to map the vision features into the inputs of LLMs. Through instruction tuning on BLIP2, InstructBLIP\citep{dai2024instructblip} gains the ability to follow the instructions on different tasks. Notably,
MiniGPT-4\cite{DBLP:journals/corr/abs-2304-10592} and MiniGPT-v2\cite{DBLP:journals/corr/abs-2310-09478} are also powerful LVLMs that
exhibit strong performance across various vision-language tasks. There are many other MLLMs such as mPLUG-Owl\citep{ye2023mplug}, Otter\citep{li2023otter} and Qwen-VL \citep{bai2023qwen}. Among all MLLMs, GPT-4V\citep{openai2023gpt4} is the most powerful one now. We select some of these MLLMs for our research. 



%



\begin{figure*}
    \centering
    \includegraphics[width=16cm]{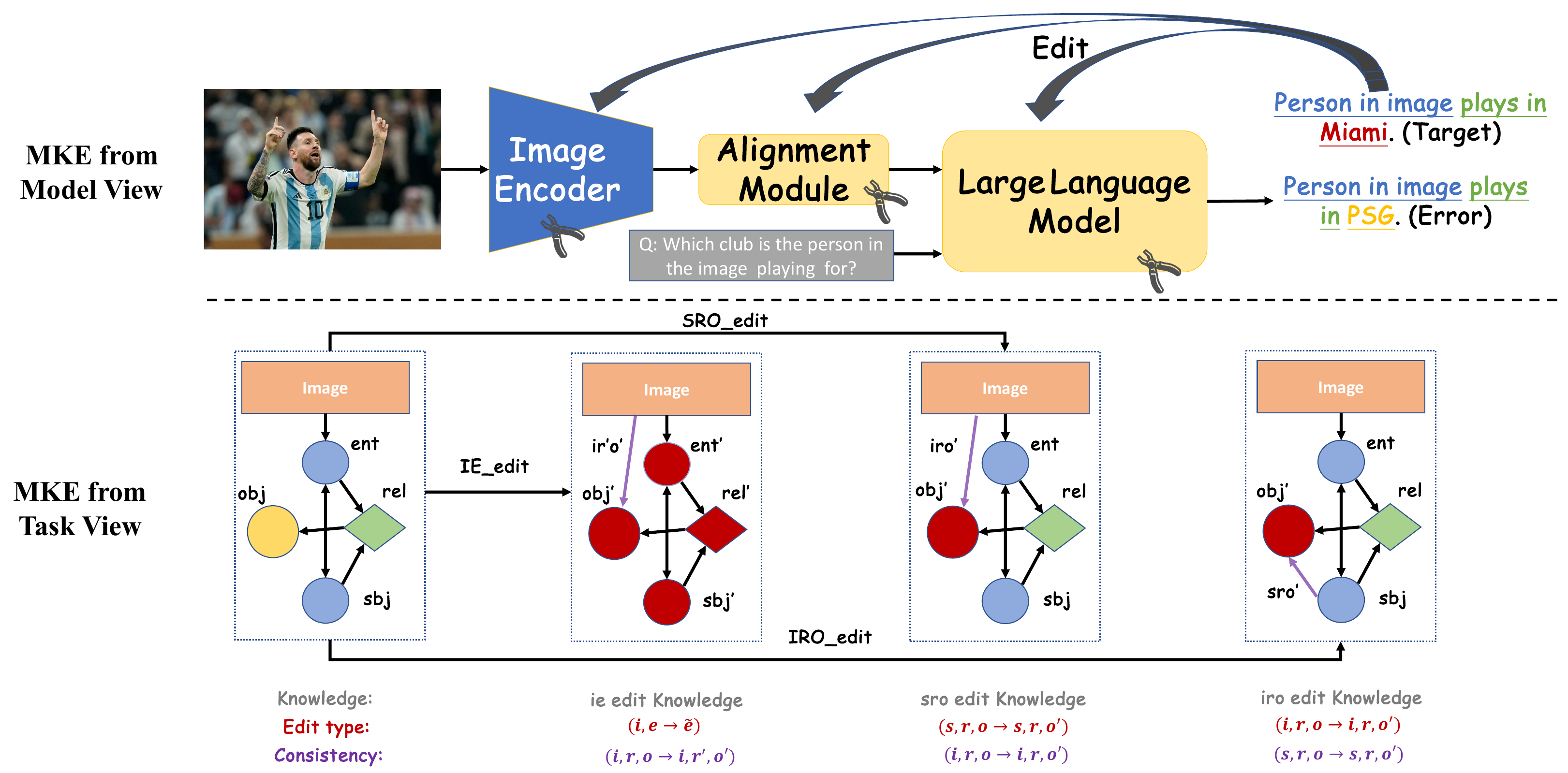}
    \caption{The upper represents editing different components of MLLMs. The bottom provides an overview of different editing formats. With an input image and its corresponding textual knowledge $(s, r, o)$, we show three different editing formats. Although the final output is the same, the edited multimodal knowledge differs when editing its visual or textual knowledge, and the consistency property is also different given different edit inputs.}
    \label{fig:edit_frame}
\vspace{-10pt}
\end{figure*}

\section{Multimodal Knowledge Editing}
\subsection{Definition of Multimodal Knowledge}

There are two types of knowledge updating scenarios, namely misrecognition and misread. The misrecognition scenario refers to the model's recognized entity from the image being incorrect and needs correcting. So we define a visual knowledge $(i, e)$ related to this scenario, where $i$ represents an image and $e$ represents the recognized entity. 

In contrast, the misread scenario focuses on the model that successfully recognizes the entity in the image but fails to provide the correct object within the context of the entity and relation. In this scenario the corresponding textual knowledge $(s, r, o)$ is related.

Therefore, we believe a piece of multimodal knowledge can be represented as a combination of visual knowledge $(i, e)$ from image recognition of an entity and textual knowledge triplet $(s, r, o)$ about the recognized entity. We finally decompose a piece of multimodal knowledge as:
\begin{equation}
    K(i, e, s, r, o) = (i, e) \times_{e=s} (s, r, o)
\label{eq:mmknowledge}
\end{equation}
\vspace{-10pt}

Further, in many cross-modal datasets, most instances represent knowledge in the final form of $(i, r, o)$ because there is no need to explicitly mention the intermediate entity $e$ (and $s$). So another combined form of multimodal knowledge can be denoted as:
\vspace{-5pt}
\begin{equation}
    (i, e) \times_{e=s} (s, r, o) = (i, r, o)
\end{equation}
\vspace{-10pt}

In summary, $(i,e), (s,r,o), (i,r,o)$ are three types of knowledge involved in multimodal knowledge editing.  However, regardless of the type of knowledge being edited, a good editing method must ensure that the consistency of multimodal knowledge is maintained after editing the corresponding type of knowledge.




\subsection{Definition of MMEdit}

We define three different edit formats, IE\_edit, SRO\_edit, and IRO\_edit.

\noindent \textbf{IE\_edit}
IE\_edit is focused on editing knowledge related to image-to-entity recognition, denoted as $(i, e)$. If we want to edit the model's recognition of an entity in an image, we input the image and modify the model's entity output for this image to a new output, which is $(i, e \rightarrow \tilde{e})$.


\noindent \textbf{SRO\_edit}
SRO\_edit is focused on editing specific textual knowledge triplets $(s, r, o)$. When we know the exact way to edit the corresponding textual knowledge tuple $(s, r, o \rightarrow \tilde{o})$, we do not need to find the corresponding multimodal data pair. Instead, we can directly use textual editing way. To ensure consistency in the input format of multimodal language models, we use a black image as visual input. Subsequent experiments in appendix \ref{sec:prelim_exps} have shown that when using questions generated from textual knowledge as input, the type of input image does not significantly impact the accuracy of the answers. In this case, the model's textual input is the same as the textual knowledge editing task.

\noindent \textbf{IRO\_edit}
In many multimodal datasets, numerous examples do not present the complete construction information about an instance of multimodal knowledge. We only possess the final multimodal data $(i, r, o)$ and may not be able to accurately decompose it into the corresponding visual knowledge and textual knowledge. 
Even though we may not explicitly identify the corresponding visual knowledge and textual knowledge, an effective method should implicitly understand and update the corresponding knowledge.

Therefore, we hope that a good multimodal knowledge editing method can maintain consistency, even when editing with the final multimodal knowledge input. Theoretically, modifying only $(i, r, o \rightarrow \tilde{o})$ should lead to consistency, whether through $(i, e \rightarrow \tilde{e})$ or $(s, r, o \rightarrow \tilde{o})$. However, there is an issue that there could be many non-unique $\tilde{e}$. Our dataset provides automatically generated reasons to determine it is a modification of $(s, r, o)$. A good editing method should automatically use the provided information to determine that the modification should be implemented on the corresponding textual knowledge triplet in IRO\_edit of our benchmark.



\subsection{Requirements of MMEdit Method}


\noindent \textbf{Consistency}
Consistency means that a piece of multimodal knowledge is answered consistently across different modalities after multimodal knowledge editing as shown in Figure \ref{fig:edit_frame}. In IE\_edit, if we modify the corresponding visual knowledge $(i, e \rightarrow \tilde{e})$, consistency means that the corresponding multimodal knowledge should also change as: 

\vspace{-15pt}
\begin{equation}
\resizebox{0.88\hsize}{!}{$
  (i, e \rightarrow \tilde{e}) \times_{\tilde{e}=\tilde{s}} (\tilde{s}, \tilde{r}, \tilde{o}) \Rightarrow (i, r, o) \rightarrow (i, \tilde{r}, \tilde{o})$
}
\label{eq:ie_consistency}
\end{equation}


In SRO\_edit, if we modify the corresponding textual knowledge $(s, r, o \rightarrow \tilde{o})$ while keeping the visual knowledge unchanged, the corresponding multimodal knowledge will also be modified to:

\vspace{-15pt}
\begin{equation}
\resizebox{0.88\hsize}{!}{
  $(i, e) \times_{e=s} (s, r, o \rightarrow \tilde{o}) \Rightarrow (i, r, o) \rightarrow (i, r, \tilde{o})$
}
\label{eq:sro_consistency}
\end{equation}
\vspace{-15pt}


In IRO\_edit, due to the reasons mentioned above, our dataset provides extra information so that when we edit multimodal knowledge $(i, r, o \rightarrow \tilde{o})$ the corresponding textual knowledge will change as follows: 

\vspace{-10pt}
\begin{equation}
\resizebox{0.88\hsize}{!}{
$(i, r, o \rightarrow \tilde{o}) \Rightarrow (i, e) \times_{e=s} (s, r, o) \rightarrow (s, r, \tilde{o})$
}
\label{eq:iro_consistency}
\end{equation}




The property of consistency imposes higher demands on the multimodal knowledge editing method, requiring that the edited knowledge remains unified across different modalities in the multimodal model.

\noindent \textbf{Reliability}
Reliability requirement of multimodal knowledge editing refers to the success rate of edits under the corresponding editing format. 

\noindent \textbf{Locality}
Locality means that multimodal editing should not affect unrelated knowledge when editing the corresponding knowledge. 

\noindent \textbf{Generality}
Generality means that after a piece of multimodal knowledge is edited, the model should not only output the edited knowledge under the exact input used for editing. It needs to provide correct edited responses under various generalizations, such as rephrased textual input or different images of the same entity.


\section{MC-MKE Benchmark Construction}

Since pure textual knowledge editing datasets are constructed from textual knowledge triplets $(s, r, o)$ and contain editing information $(s, r, o \rightarrow \tilde{o})$, we opt for using the textual knowledge editing dataset MQuAKE \cite{zhong2023mquake} as the starting point to construct our multimodal knowledge editing dataset MC-MKE. MQuAKE, as a text knowledge editing dataset, contains knowledge triplets, related editing information and questions as test input. Each instance in MQuAKE corresponds to a textual knowledge triplet and its textual editing information.


\subsection{Data Selection} 
\label{sec:data_select}
Unlike previous editing datasets, we performed filtering in three directions step by step on the original MQuAKE dataset $D_{raw}$ to achieve a high-quality dataset.

We first using a completely black image paired with textual questions in MQuAKE to ask the MLLMs to filter the data to obtain $D_{filter_1}$. From $D_{filter_1}$, we obtain related images of subjects in the questions from Google, and then apply questions which replace subject with its category to ask MLLMs together with corresponding image to obtain filtered data $D_{filter_2}$. Finally, we transform the textual questions into multimodal questions and apply these question to obtain the final multimodal knowledge editing source dataset $D_{orig}$. We utilize some of the filtered-out data to construct a training set for methods requiring training dataset to adjust parameters (such as SERAC). More details about data selection and generation quality assessment can be found in Appendix \ref{sec:data_details}.

%



\begin{table*}[htbp]
\centering
\small
\setlength{\tabcolsep}{5mm}{
\begin{tabular}{c c c c c c c} 
%


\toprule

Model          & Method         & $\text{Score}_R$ & $\text{Score}_L$ & $\text{Score}^T_G$ & $\text{Score}^M_G$ & $\text{Score}_C$ \\
\midrule 
\multirow{6}*{InstructBLIP}   & FT(Vision)      &  89.57           &  0.34        &     24.10         &      90.30        &    38.07         \\
               & FT(LLM)        & \textbf{98.48}            &   0.03       &    78.04          &    \textbf{96.41}          & 9.09            \\
               & MEND(Vision)    & 32.39 & \textbf{93.15} & 29.73 & 23.43 & 18.37 \\
               & MEND(LLM)      & 88.58 & 53.23 & \textbf{86.49} & 85.21 & 9.46\\
               & IKE            & 68.26            &     /        & 76.33             &      /       &    \textbf{49.05}         \\
               & SERAC          & \textbf{98.48}            & 87.65        & 68.41         &  \textbf{96.41}    & 9.09            \\
\midrule 
\multirow{6}*{MiniGPT-v2}      
    & FT(Vision) & \textbf{98.04} & 66.43 & \textbf{98.13} & \textbf{91.52} & 16.67 \\ 
               & FT(LLM) & 95.76 & 0.59 & 93.41 & 91.48 & 8.71 \\
               & MEND(Vision)    & 7.57 & 56.73 & 6.17 & 5.69 & 11.36\\
               & MEND(LLM)      &     26.52        &    67.34      &     29.19       &      20.17     &     4.54       \\
               & IKE            &    47.61            &    /     &  25.24     &      /       & \textbf{60.60}                \\
               & SERAC               &    95.76         &    \textbf{83.85} & 81.48  &  91.48       &    8.71       \\

\bottomrule
\end{tabular}
}
\caption{Experimental results on IE\_edit data for four editing methods editing two different model components on two MLLMs. The highest value is highlighted in \textbf{bold}.}
\label{table:ie_edit}
\vspace{-10pt}
\end{table*}


\subsection{Dataset Construction}

\paragraph{Editing Dataset Construction}

For multimodal knowledge in our filtered multimodal knowledge source dataset $D_{orig}$, we sequentially construct editing data under different editing formats. For IE\_edit, our editing inputs consist of images and automatically generated questions. We choose to use an entity $\tilde{e}$ of the same category as the entity $e$ as the editing target. For SRO\_edit, our editing inputs consist of textual questions, with the editing target being the corresponding new knowledge $\tilde{o}$ given in MQuAKE dataset. We require that $\tilde{o}$ is of the same entity category as $o$. For IRO\_edit, our editing input is constructed based on the input from SRO\_edit, combined with entity types and templates. The target $\tilde{o}$ is chosen from the corresponding data in the SRO\_edit editing dataset.  more strict requirements can be seen in appendix \ref{sec:data_details}.



\noindent \textbf{Reliability Dataset Construction}
Our Reliability metric is calculated as shown in the following formula. $D_e$ is the editing dataset corresponding to the editing format.  For each piece of multimodal knowledge $k = (i, e) \times (s, r, o)$ in $D_e$, $\tilde{k}$ is the corresponding edited knowledge. $p_r$ is the multimodal input used for testing the Reliability of the corresponding editing format. $t_r$ is the target reliability output after knowledge editing. $F$ is the multimodal model, and $\theta_{k\tilde{k}}$ represents the parameters of the model after editing a multimodal knowledge $k \rightarrow \tilde{k}$.
\vspace{-5pt}
\begin{equation}
    \text{Score}_{R} = \mathbb{E}_{(k, \tilde{k}, p_{r}, t_r) \sim D_{e}} \left[ \mathbbm{1}_{F(p_{r};\theta_{k\tilde{k}}) = t_r} \right]
\label{eq:s_r}
\end{equation}
\vspace{-10pt}


\noindent \textbf{Consistency Dataset Construction}
The construction of our Consistency knowledge editing dataset varies depending on the different editing formats.
In IE\_edit, consistency is defined as Eq (\ref{eq:ie_consistency}). Therefore, after editing visual knowledge, we construct the input $p_c$ corresponding to the multimodal knowledge. The edited model should output the corresponding $\tilde{o}$ for this input to ensure consistency. In SRO\_edit, consistency is defined as Eq (\ref{eq:sro_consistency}). We will edit the corresponding textual knowledge triplet, and then construct the multimodal input $p_c$ for multimodal knowledge to test whether the edited model can provide a consistent edited answer $\tilde{o}$ given input $p_c$. In IRO\_edit, consistency is defined as Eq (\ref{eq:iro_consistency}). For each piece of multimodal knowledge, we find its corresponding textual knowledge. After editing the multimodal knowledge, we will analyze whether the corresponding textual knowledge provides a consistent response. The consistency score is shown in the following formula. $p_c$ is the multimodal input, $\theta_{k\tilde{k}}$ is the edited parameters, $t_c$ is the corresponding consistency output in different editing format. Others are the same as Eq (\ref{eq:s_r}).
\vspace{-10pt}
\begin{equation}
    \text{Score}_C = \mathbb{E}_{(k, \tilde{k}, p_c, t_c) \sim D_e} \left[ \mathbbm{1}_{F(p_c;\theta_{k\tilde{k}}) = t_c} \right]
\end{equation}
\vspace{-10pt}



\noindent \textbf{Locality Dataset Construction}
In the edited datasets for the three editing formats, we used data unrelated to the current editing knowledge but of the same editing format as locality data. In IE\_edit, we randomly selected visual information ($i_{loc}$, $e_{loc}$) different from the current entity in $D_{orig}$ as locality data. In SRO\_edit, we randomly selected data $(s_{loc}, r_{loc}, o_{loc})$ different from the current textual knowledge triplet $(s, r, o)$ in $D_{orig}$ as locality data. In IRO\_edit, we randomly selected multimodal knowledge $(i, e) \times_{e=s} (s, r, o)$ where $i, e, s, r$, and $o$ are all different in $D_{orig}$ to form locality data $(i_{loc}, e_{loc}) \times_{e_{loc}=s_{loc}} (s_{loc}, r_{loc}, o_{loc})$.

The locality score is shown in the following formula. $p_l$ is the multimodal input, $\theta_{k\tilde{k}}$ is the edited parameters, $t_l$ is the corresponding locality output in different editing format.
\vspace{-10pt}
\begin{equation}
    \text{Score}_L = \mathbb{E}_{(k, \tilde{k}, p_l) \sim D_e} \left[  \mathbbm{1}_{F(p_l;\theta_{k\tilde{k}}) = t_l}  \right]
\end{equation}

\begin{table}[tb]
\centering
\small
\setlength{\tabcolsep}{3pt}
\renewcommand{\arraystretch}{1.25}
\resizebox{1\linewidth}{!}{
\begin{tabular}{cccccc}
    \toprule
    & Edit format & IE\_edit & SRO\_edit & IRO\_edit & All\\
    \midrule
    \multirow{7}*{\rotatebox{90}{Train}} & \#Data & 3544 & 5968 & 5968 & 15480 \\
    & \#Relation & 37 & 30 & 30 & 37 \\
    & \#Entity & 3544 & 5230 & 5230 & 5407 \\
    & \#Alias(avg.) & 14.18 & 13.62 & 13.62 & 13.75 \\
    & \#Image & 21264 & - & 20790 & 22134 \\
    & \#Category & 142 & 342 & 342 & 343 \\
    & \#Input Samples  & 28352 & 35808 & 47744   & 111904 \\
    \midrule
    \multirow{7}*{\rotatebox{90}{Test}} & \#Data & 920 & 982 & 982 & 2884 \\
    & \#Relation & 28 & 30 & 30 & 30 \\
    & \#Entity & 810 & 1041 & 1041 & 1424\\
    & \#Alias(avg.) & 20.46 & 17.02 & 17.02 & 18.11\\
    & \#Image & 2358 & - & 1311 & 2550\\
    & \#Category & 49 & 76 & 76 & 76\\
    & \#Input Samples &  15640   & 11784  &  16694   &  44118 \\
    \bottomrule
\end{tabular}}
\caption{The statistic of different subsets of MC-MKE. \lstinline|\#Entity| refers to the total number of entities appeared including $s, o$ and $e$. \lstinline|\#Alias| refers to the number of answer aliases. \lstinline|\#Image| in Test refers to the number of filtered images in $D_{orig}$}
\label{tab:stat}
\vspace{-10pt}
\end{table}

\begin{table*}[ht]
\centering
\small
\setlength{\tabcolsep}{7mm}{
\begin{tabular}{c c c c c c c} 
\toprule

Model          & Method         & $\text{Score}_R$ & $\text{Score}_L$ & $\text{Score}^T_G$  & $\text{Score}_C$ \\
\midrule 
\multirow{6}*{InstructBLIP}   & FT(Vision)      &  91.75      &    4.23  &     17.84     &    87.57         \\
               & FT(LLM)        &  \textbf{99.49}      & 3.95 & \textbf{79.59} & \textbf{90.43}  \\
               & MEND(Vision)    & 13.64 & \textbf{95.03} & 10.00 & 3.86 \\
               & MEND(LLM)      & 66.49 & 79.34 & 72.85 & 55.90\\
               & IKE            & 81.06         & 94.18         & 55.87  &  73.73           \\
               & SERAC          & \textbf{99.49}         & 89.53         & 65.05  &  \textbf{90.43}           \\ 
\midrule 
\multirow{6}*{MiniGPT-v2}      
           & FT(Vision) & \textbf{98.78} & 24.81 & \textbf{97.68} & 31.67 \\ 
               & FT(LLM) & 97.35 & 2.01 & 93.73 & \textbf{91.24} \\
               & MEND(Vision)    & 4.37 & \textbf{93.50} & 3.29 & 2.74\\ 
               & MEND(LLM)     & 2.85 & 76.96 & 2.62 & 3.25 \\ 
               & IKE       & 30.55 & 91.26 & 24.83 & 21.18 \\   
               & SERAC           & 97.35 & 91.43 & 75.05 & \textbf{91.24} \\

\bottomrule
\end{tabular}}
\caption{Experimental results on SRO\_edit data for four editing methods editing two different model components on two MLLMs. The highest value is highlighted in \textbf{bold}.}
\label{table:sro_edit}
\end{table*}

\noindent \textbf{Generality Dataset Construction}
For the three forms of multimodal knowledge editing IE\_edit, SRO\_edit, and IRO\_edit, we constructed corresponding generalization test datasets from both image and text perspectives. For the image generalization dataset, we used CLIP to process the images previously crawled from the web. Then, we calculated the relevance between the images and entities using the CLIP model and selected the top 5 most relevant images as the test images for entity image generalization. For the text generalization dataset, we use ChatGPT to rewrite 5 variations of the textual input to serve as the test inputs for text generalization. The prompts and quality assessment can be seen in appendix \ref{sec:data_details}.

The generality score is shown in the following formula. $p^T_g$, $p^M_g$is the multimodal input for text, and image generalization testing, respectively. $\theta_{k\tilde{k}}$ is the edited parameters.
$t^T_g$, $t^M_g$ are the corresponding text, and image generality output, respectively, in different editing formats.
\vspace{-10pt}
\begin{equation}
    \text{Score}^T_G = \mathbb{E}_{(k, \tilde{k}, p^T_g, t^T_g) \sim D_e} \left[  \mathbbm{1}_{F(p^T_g;\theta_{k\tilde{k}}) = t^T_g}  \right]
\end{equation}
\vspace{-10pt}
\begin{equation}
    \text{Score}^M_G = \mathbb{E}_{(k, \tilde{k}, p^M_g, t^M_g) \sim D_e} \left[  \mathbbm{1}_{F(p^M_g;\theta_{k\tilde{k}}) = t^M_g}  \right]
\end{equation}
\vspace{-10pt}

Construction details about multimodal input $p$ and corresponding $t$ can be seen in appendix \ref{sec:data_details}.









\noindent \textbf{Benchmark statistics} 
We create MC-MKE consisting of a training set with 111904 samples and a test set with 44118 samples. Methods such as SERAC can apply training set to adjust their configuration. The test set consists of a total of 2884 pieces of knowledge across three different edit formats. The associated knowledge involves a large number of entities and relations, indicating the diversity of MC-MKE. It also has an average of $18.11$ answer aliases per sample, significantly reducing misjudgments of the exact match metrics. More details about dataset statistics are presented in Table \ref{tab:stat}.

\section{Experiments}

\subsection{MMEdit Methods}


There have been many language knowledge editing methods, while multi-modality knowledge editing methods have not been fully explored. Therefore, we select the following representative editing methods including Finetuning, MEND \citep{mitchell2022fast}, IKE \citep{zheng2023edit} and SERAC \citep{pmlr-v162-mitchell22a} in single-modal knowledge editing following previous setting\citep{cheng2024edit}. More information and implementation details of these editing methods can be seen in Appendix \ref{sec:experimental details}.

\begin{table*}[ht]
\centering
\small
\setlength{\tabcolsep}{5mm}{
\begin{tabular}{c c c c c c c} 
\toprule

Model          & Method         & $\text{Score}_R$ & $\text{Score}_L$ & $\text{Score}^T_G$ & $\text{Score}^M_G$ & $\text{Score}_C$ \\
\midrule 
\multirow{6}*{InstructBLIP}   & FT(Vision)      &   84.83          &  2.75         & 34.25             &   85.07           &   76.37          \\
               & FT(LLM)    &     \textbf{91.65}           &  4.85        & 81.87              &  \textbf{91.47}      & \textbf{86.46}       \\
               & MEND(Vision)    & 24.13 & 85.88 & 33.11 & 19.20 & 5.49\\
               & MEND(LLM)      & 70.57 & 64.78 & \textbf{86.00} & 72.05 & 50.50 \\
               & IKE            & 71.59 &    /  & 82.83      &       /       &  48.17          \\
               & SERAC          & \textbf{91.65}      & \textbf{99.06}    &  26.01    &  \textbf{91.47} &  \textbf{86.46}         \\ 
\midrule 
\multirow{6}*{MiniGPT-v2}      
            & FT(Vision) & \textbf{98.98} & 73.71 & \textbf{98.78} & \textbf{93.32} & 24.13 \\
               & FT(LLM) & 88.49 & 2.04 & 86.99 & 87.25 & \textbf{84.32} \\
               & MEND(Vision)    & 6.21 & 76.00 & 5.45 & 4.52 & 2.13\\
               & MEND(LLM)      & 34.21 & 67.31 & 43.91 & 25.49 & 6.72 \\
               & IKE            & 62.73 &    /      & 62.48 &       /      & 21.49 \\
               & SERAC          & 88.49 & \textbf{97.25} & 26.92 & 87.25 & \textbf{84.32} \\

\bottomrule
\end{tabular}}
\caption{Experimental results on IRO\_edit data for four editing methods editing two different model components on two MLLMs. The highest value is highlighted in \textbf{bold}.}
\label{table:iro_edit}
\end{table*}

\subsection{Results \& Analysis}

\paragraph{Consistency On Different Editing Formats}
In SRO\_edit and IRO\_edit, the output of their corresponding Consistency output matches the required edited output, with only the input information being different. 
In these two editing formats, high Consistency without high Locality may come from overfitting. Thus, to accurately assess the Consistency property, we need to analyze the IE\_edit format as well, where the Consistency output $\tilde{o}$ is different from edited output $\tilde{e}$. 

According to Table \ref{table:ie_edit}, the FT(Vision) maintains high Consistency with InstructBLIP, indicating that the FT(Vision) is not solely overfitting to obtain Consistency in IE\_edit dataset. Only when a method achieves high Consistency across all three editing formats can its Consistency property be considered trustworthy.  Overall, IKE shows good Consistency while maintaining a certain degree of Locality. However, even IKE shows unsatisfactory consistency performance on the IE\_edit dataset, and its performance on MiniGPT-v2 on the other two datasets is even worse. This indicates that the current tested methods may ignore Consistency during development, resulting in their inability to maintain high consistency across all datasets.

\paragraph{Pros and Cons of Different Editing Methods}
The Finetuning method is characterized by high Reliability. It also demonstrates good Generality and Consistency when editing the LLM part in SRO\_edit and IRO\_edit. However, its Locality is not satisfying, suggesting it has a significant impact on unrelated knowledge. 

MEND employs a meta-learning approach to adjust the model's parameters while minimizing effects on unrelated knowledge. As shown from the results, MEND bears a much lower Reliability but higher Locality than Finetuning. The meta-learning approach, while preventing the modification of knowledge irrelevant to the model, also reduces the accuracy of editing relevant knowledge. Moreover, the performance of meta-learning is highly unstable across different types of data and models, according to results in Table \ref{table:sro_edit} and Table \ref{table:iro_edit}. 

As for IKE, since many MLLMs do not support in-context learning for image inputs, we do not test it for Locality and M-Generality in IE\_edit and IRO\_edit. IKE achieving good Locality in SRO\_edit. However, IKE relies on in-context learning, which is inherently sensitive to prompts. Different types of editing formats require different prompts, and different models have varied sensitivity degrees to prompts, resulting in significant fluctuations in all metrics of IKE. Overall, IKE performs better on InstructBLIP. In IE\_edit, it also shows high Consistency on InstructBLIP, achieving the highest Consistency. indicating that the model can infer the edited multimodal knowledge $(i, \tilde{r}, \tilde{o})$ based on context and the image.

SERAC applies a classifier to choose whether to use the original model or the counterfactual model. SERAC achieves high levels of Reliability and Locality, as well as Generality in most cases. However, its performance relies heavily on the classifier performance, whether the classifier can correctly identify the appropriate model for the given input. Although SERAC obtains good Consistency with good Locality in SRO\_edit and IRO\_edit datasets, its Consistency in IE\_edit is still low. While the classifier can sometimes effectively distinguish between inputs related to edited knowledge and those that are not, it still cannot directly improve Consistency. Even if the classifier identifies the need to use the counterfactual model to answer questions in the Consistency test, the ability to respond to the Consistency test still depends on the counterfactual model itself.



We observe the results across all editing formats and models, no existing editing method perfectly meets all editing requirements. Most methods have problems with consistency, according to Table \ref{table:ie_edit}, \ref{table:sro_edit} and \ref{table:iro_edit}.



\paragraph{Editing Different Components}
\citet{cheng2024edit} mentioned the visual module is harder to edit compared to the text module. Based on our experimental results, this point holds true in some cases, but in other instances, editing the visual part may yield better results in certain aspects.
For MEND, meta-learning requires predicting network changes corresponding to the knowledge edits, and editing the visual module to output the edited knowledge is more challenging. As a result, in most cases, using MEND(Vision) tends to result in lower reliability. 



While the MEND approach does help prevent the modification of irrelevant knowledge to some extent, editing the LLM module with MEND still often achieves lower locality as shown in Table \ref{table:sro_edit} and \ref{table:iro_edit}. Furthermore, in the IE\_edit dataset, MEND achieves higher consistency when editing the visual module. Across the three datasets, FT(Vision) often achieves reliability similar to FT(LLM). On MiniGPT-v2, FT(Vision) results in higher locality. When editing MiniGPT-v2, although FT(LLM) tends to yield higher consistency on the SRO\_edit and IRO\_edit datasets, it comes with low locality. The higher consistency brought by FT(LLM) in SRO\_edit and IRO\_edit datasets may be due to overfitting to the edited output. This is evidenced by the IE\_edit dataset, where the consistency output is different from the edited output. In IE\_edit dataset, FT(LLM) yields worse consistency than FT(Vision). What's more, in IE\_edit dataset, FT(Vision) achieves the highest Consistency among all parameter-based methods, which may indicate that editing the Vision part may be better when editing visual knowledge.

\section{Conclusion}
We refine the definition of multimodal knowledge and introduce a new benchmark MC-MKE. We conduct experiments to analyze the effectiveness of several multimodal knowledge editing methods across different models, editing formats, and components. We find that these methods have limitations, and cannot achieve perfect performance on different editing formats. To maintain consistency, it may be better to edit the model components corresponding to the specific knowledge part.

\section*{Limitations}
The main limitations of our work are related to limited knowledge editing methods and multimodal large language models. We only provide results on MLLMs with 7B checkpoint. We were unable to test larger checkpoints, due to resource constraints.
As we study the latest MLLMs on four knowledge editing methods which have not been discussed in prior work, we need to implement them from scratch. We end up implement four knowledge editing methods, Finetuning, MEND, IKE and SERAC.


\section*{Ethical Considerations}
MC-MKE: is a synthetic dataset constructed by randomly modifying the factual knowledge triplets, rather than being crafted by humans. The data samples could accidentally involve context which is toxic or offensive in nature. ChatGPT is used for data annotation and assisting writing.


\bibliography{acl_latex}

\appendix


\section{Pre-experiments}
\label{sec:prelim_exps}


SRO\_edit focuses on editing a textual knowledge triplets $(s,r,o)$, inherently requiring no additional visual inputs. But to align with the standard input format of MLLMs, we input a black image as the visual placeholder. In this section, we present a preliminary experiment to explore different choices of the input visual images including black images, white images and random noise. The accuracy of InstructBLIP with these three types of images on SRO\_edit are 95.11, 96.53 and 94.70 respectively. It is shown that these uninformative images barely have influence on the results.



\section{Experiment Details}
\label{sec:experimental details}
\paragraph{Finetuning Details}
Finetuning is one of the most widely used and apparent methods for improving or modifying the abilities of pre-trained models and is also generally used as a baseline for knowledge editing. Since one can select the model component to finetune, it is natural to explore the differences between finetuning different model components. We focus on finetuning two parts: the alignment module and the LLM component of an MLLM. For the LLM component, we only finetune the last layer. 
We list the hyper-parameters used for finetuning in Table \ref{tab:app-hyper-sft}. MiniGPT-v2 and InstructBLIP share the same hyper-parameters.

\begin{table}[htbp]
    \centering
    \setlength{\tabcolsep}{8mm}{
    \begin{tabular}{l|c}
    \hline
        Learning Rate & 5e-4 \\
        Steps & 16 \\
        Optimizer & AdamW \\
        Weight Decay & 0.05 \\
    \hline
    \end{tabular}
    }
    \caption{Hyper-Parameters used for finetuning.}
    \label{tab:app-hyper-sft}
\end{table}

\paragraph{MEND Details}
Model Editor Networks with Gradient Decomposition (MEND) \citep{mitchell2022fast} is an editor network mapping a single desired input-output knowledge pair to the corresponding parameter update of the original model. Specifically, the input-output knowledge pair provides a standard fine-tuning gradient as a starting point for editing updates. Then MEND directly transforms the gradient to a better parameter update ensuring both generality and locality. Training process of MEND requires additional training data specific to the underlying model. Following \cite{mitchell2022fast}, we construct an edit dataset and a locality dataset for both InstructBLIP and MiniGPT-v2. We leverage the data filtered in Section \ref{sec:data_select} as the edit dataset, sharing identical distribution with MC-MKE. Since both InstructBLIP and MiniGPT-v2 leverage MS COCO\cite{lin2015microsoft} for pretraining, we include it as the locality training dataset. We search for three important hyper-parameters $c_{loc}$, $c_{edit}$ and learning rate on each experimental setting for ten times. We found that MEND is very sensitive to hyperparameters, especially when the target module is small (e.g. the MEND(Vision) setting in our main experiment).

\paragraph{IKE Details}
In-Context Knowledge Editing (IKE)\citep{zheng2023edit} enables knowledge editing by incorporating demonstration examples within the input data to update and acquire new factual knowledge without the requirement of further training. Considering the limitation on the number of input images, we choose to implement the zero-shot version of IKE. 

\paragraph{SERAC Details}
SERAC\citep{pmlr-v162-mitchell22a} proposes a memory-based editing approach. The approach consists of a classifier and a counterfactual model. The classifier chooses whether to use the counterfactual model or not based on the relation between the given input and edit memory. 

Since our tasks are multimodal, we use a neural network trained on the training set as the classifier. The neural network consists of a CLIP feature extraction layer and an MLP classification layer. We set the learning rate of the classification layer to 0.0005. Since consistency requires the model to have reasoning abilities, we opted to continue using the large model as the counterfactual model. Specifically, we employ a large model with its LLM part fine-tuned on the edited knowledge as the counterfactual model. 

\paragraph{MLLMs Details}
InstructBLIP is a multimodal large language model that consists of three modules. Its multimodal alignment module consists of a Qformer structure and a linear layer network to connect its vision and large language model module. We use InstructBLIP equipped with Vicuna-7B \cite{vicuna2023}.

MiniGPT-v2 utilizes a linear projection layer as an alignment module to map visual features to LLM feature space. Compared with InstructBLIP, MiniGPT-v2 has a smaller alignment module but still more input visual features. We use MiniGPT-v2 equipped with Llama-2-Chat-7B \cite{touvron2023llama}. 


\section{Data Details}
\label{sec:data_details}

\paragraph{Data Selection Details}
We filter the data using a completely black image paired with questions in MQuAKE dataset. We selected data that our MLLMs could correctly answer. This step ensures that all the edited knowledge is originally known by the model to make sure we are ``editing" instead of ``learning". The filtered dataset is referred to as $D_{filter_1}$. 


From $D_{filter_1}$, we obtain related images from Google, of the subject $s$ in the textual knowledge triplets $(s, r, o)$. We then used ChatGPT to generate fine-grained entity categories for these subjects and construct image queries using specific templates. If the subject in the image could be correctly recognized by all MLLMs, the data is then retained. This step ensures that all entities in our dataset are known by the models. This constitutes the dataset $D_{filter_2}$.

Finally, we replaced the subject in the questions with “the \{category\} in the picture”, seen in Appendix \ref{sec:prompts}. If the combined question can be correctly answered by all models, the data is then retained. This step ensures the original multimodal knowledge consistency. The final retained multimodal knowledge constitutes our knowledge editing source dataset $D_{orig}$. 

\noindent \textbf{Entity Alias}
To facilitate entity evaluation, we collect alias of entities for all answers from the original dataset $D_{raw}$. However, since we will edit some of the subject entities, we also used alias data from Wiki as a supplement to construct the final entity alias library. All of our matching is performed with entities and their corresponding aliases.

\noindent \textbf{Edit input Construction Details}
We choose to use an entity $\tilde{e}$ of the same category as the entity $e$ and we require that the corresponding textual knowledge triplet $(\tilde{s}, \tilde{r}, \tilde{o})$, which $\tilde{s}=\tilde{e}$ exists in $D_{filter_1}$.

\noindent \textbf{Locality Construction Details}
We ensure that these selected entities differ from those of the current knowledge.
Formally, the knowledge $K_{loc}(i',e',s',r',o')$ for locality test of knowledge $K(i,e,s,r,o)$ must satisfy the condition $i'\neq i, e'\neq e, s'\neq s, r'\neq r,o'\neq o$.
We randomly sample five pieces of knowledge to serve as the locality test data.

\paragraph{Entity Category Generation Evaluation}
We employ ChatGPT to generate the category of a given entity. To verify the quality of categories generated by ChatGPT, we randomly sampled 200 items and invited two annotators to independently verify whether the entities mentioned in these items matched their respective categories. 
The average agreement between the annotators was 98\%, with a consistency rate of 97\%, indicating that the generated entity categories are highly reliable. An example of a generated entity category is: "google" : "company". 

\paragraph{Training set Construction}
Except for not undergoing the original problem filtering, the construction of the train data is similar to that of the test set. We utilize some of the filtered data to construct training set. For the filtered data which are not in $D_{orig}$, we directly apply the question in MQuAKE dataset as text generality textual input, and use the images from google as image generality visual input. 

\paragraph{Rephrase Generation Evaluation}
We employ ChatGPT to generate Generality data. To verify the quality of rephrases generated by ChatGPT, we randomly sampled 100 items each associated with 4 paraphrased sentences and asked two annotators to independently assess the quality of each paraphrased sentence, marking them as 0 for bad quality and 1 for good quality. 
The average scores for the 400 paraphrase results were 0.9675, respectively, with an agreement of 98\%, demonstrating that the quality of our paraphrases is sufficiently reliable. An example of paraphrased sentence is: Origin : "Who performed Folsom Prison Blues?" Rephrase : "Who was the performer of Folsom Prison Blues?"

\section{Prompts}
\label{sec:prompts}
    We designed specific prompts and instructions for GPT-3.5-turbo-16k to rephrase the textual input for the text generalization dataset and generate fine-grained entity types, as shown in Table \ref{tab:prompt_rephrase} and Table \ref{tab:prompt_entity}, respectively.
    
    We provide editing and testing inputs of different types of multimodal knowledge editing  in Table \ref{tab:input_prompt}, Table \ref{tab:sro} and Table \ref{tab:iro}.

\begin{table}
    \centering\small
    \begin{tabular}{p{7.2cm}}
    \toprule
    \textbf{Prompts and Instructions} \\
    \midrule 
    You are a helpful assistant. \\
    Please rephrase the following original text with 10 different and diverse expressions, maintaining exactly the same meanings. \\
    Note that you must not add any additional information and not delete or lose any information of the original text. \\
     \\
    Original Text: \\
    \{source\} \\
     \\
    5 Rephrased Texts: \\
    \bottomrule
    \end{tabular}
    \caption{Prompts and instructions used for rephrasing the textual input for the text generalization dataset.}
    \label{tab:prompt_rephrase}
\end{table}

\begin{table}
    \centering\small
    \begin{tabular}{p{7.2cm}}
    \toprule
    \textbf{Prompts and Instructions} \\
    \midrule 
    You are a powerful fine-grained entity category generator. User will give the name of entity, and you will help answer the fine-grained categoty of the entity. The answer is the categoty only. \\
    There are some examples: Given entity Cameroon, a possible answer should be "country". \\
    Given entity David Beckham, a possible answer should be "person". \\
    Given entity The Great Gatsby, a possible answer should be "book". \\
    Given entity Producers' Showcase, a possible answer should be "TV show". \\
    Given entity Lady Madonna, a possible answer should be "song". \\
    Given entity Cox Enterprises, a possible answer should be "company". \\
    The given entity is \{\}, a possible answer is: \\
    \bottomrule
    \end{tabular}
    \caption{Prompts and instructions used for generating fine-grained entity types.}
    \label{tab:prompt_entity}
\end{table}

\begin{table}
    \centering
    \small
    \setlength{\tabcolsep}{4mm}{
    \begin{tabular}{m{10pt}m{0.5in}m{1.5in}}
    \toprule
    \centering
    Input & \makecell[c]{Visual \\Inputs} & Textual Inputs \\
    \midrule
    \centering{Edit\\input} & 
    \includegraphics[width=0.5in]{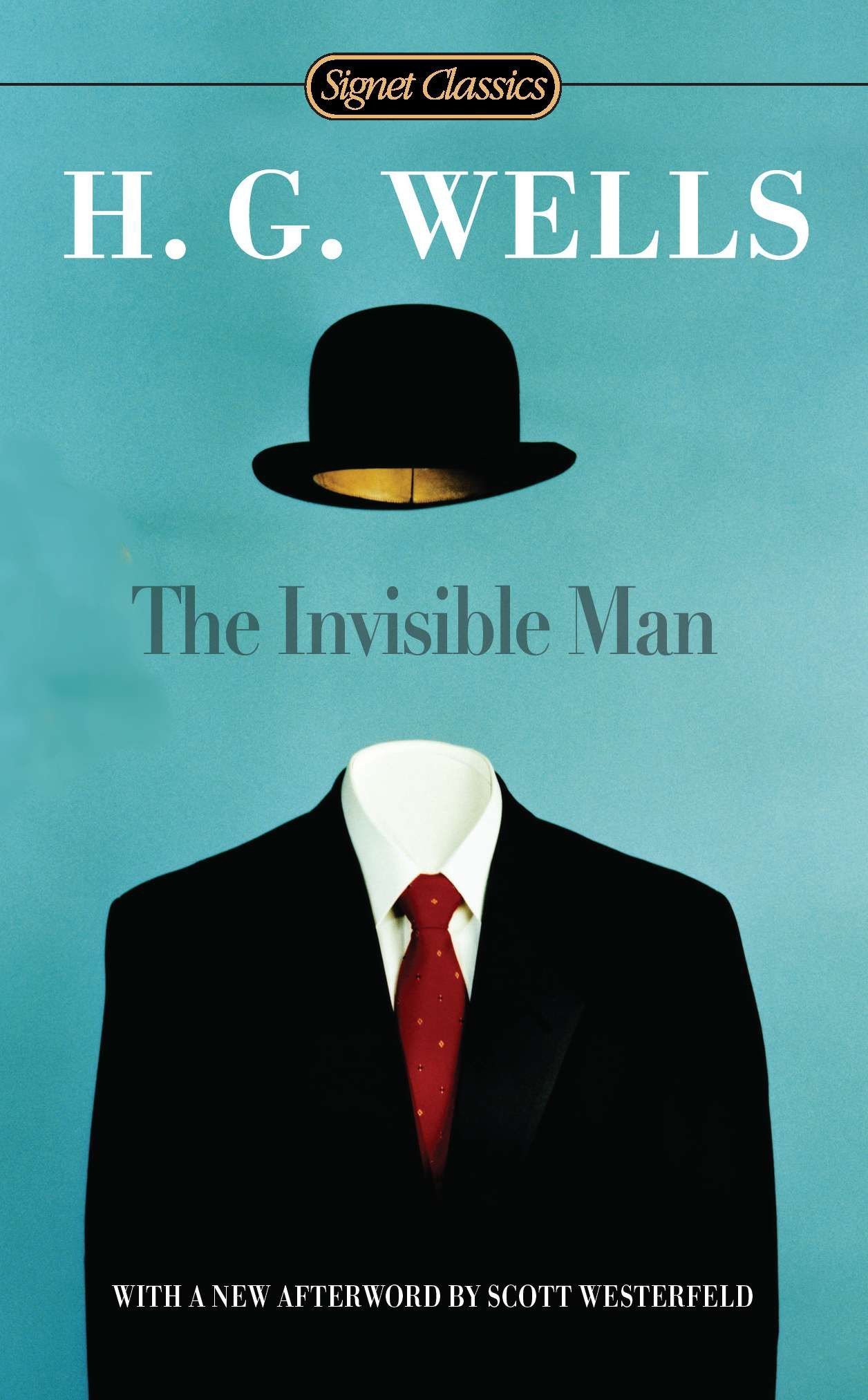} &      Question: The book in the picture is \newline
    $\tilde{e}$: The Pilgrim's Progress\newline
    \\
    \midrule
    \centering{$p_r$} & 
    \includegraphics[width=0.5in]{figures/image1.jpg} &      Question: The book in the picture is \newline
    $t_r$: The Pilgrim's Progress\newline
    Alias: Pilgrim's Progress, Land of Beulah, ...\\
    \midrule
    \centering{$p_c$} & 
    \includegraphics[width=0.5in]{figures/image1.jpg} &      Question: The book in the picture was written in the language of \newline
    $t_c$: English\newline
    Alias: en, eng, English language, ...\\
    \midrule
    \centering{$p_l$} & 
    \includegraphics[width=0.5in]{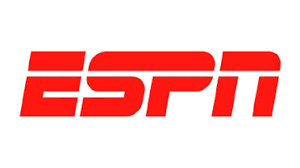} \newline
     &      
    Question: Which TV channel is shown in the picture? \newline
    $t_l$: ESPN\newline
    Alias: Entertainment and Sports Programming Network 
    \newline
     \\
    \midrule
    \centering{$p^M_g$} & 
    \includegraphics[width=0.5in]{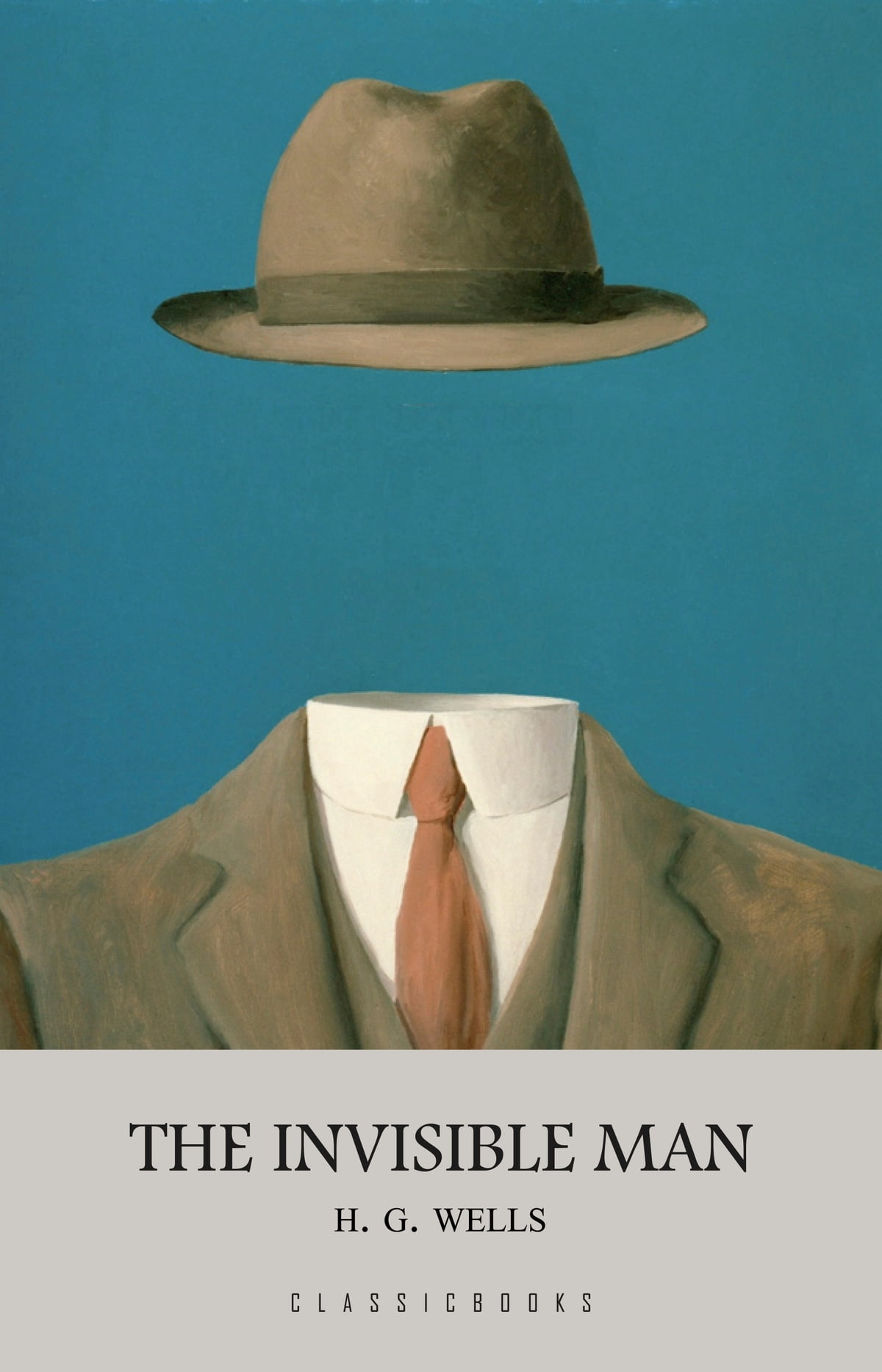} \newline
     &      
    Question: The book in the picture is \newline
    $t^M_g$: The Pilgrim's Progress\newline
    Alias: Pilgrim's Progress, Land of Beulah, ...\newline
     \\
    \midrule
    \centering{$p^T_g$} & 
    \includegraphics[width=0.5in]{figures/image1.jpg} \newline
     &      
    Question: Which book is shown in the picture? \newline
    $t^T_g$: The Pilgrim's Progress\newline
    Alias: Pilgrim's Progress, Land of Beulah, ...\newline
     \\
    \bottomrule
    
    \end{tabular}
    }
    \caption{IE\_edit multimodal input examples.}
    \label{tab:input_prompt}
\end{table}

\begin{table}
    \centering\small
    \setlength{\tabcolsep}{4mm}{
    \begin{tabular}{m{10pt}m{0.5in}m{1.5in}}
    \toprule
    \centering
    Input & \makecell[c]{Visual \\Inputs} & Textual Inputs \\
    \midrule
    \centering{Edit\\input} & 
    \centering{/}
    &  Question: Invisible Man was written in the language of \newline
    $\tilde{o}$: Sanskrit\newline
    \\
    \midrule
    \centering{$p_r$} & 
    \centering{/}
    &  Question: Invisible Man was written in the language of \newline
    $t_r$: Sanskrit\newline
    Alias: Samskrta, Samskrtam, Sanskrit language, ...\\
    \midrule
    \centering{$p_c$} & 
    \includegraphics[width=0.5in]{figures/image1.jpg} &      Question: The book in the picture was written in the language of \newline
    $t_c$: Sanskrit\newline
    Alias: Samskrta, Samskrtam, Sanskrit language, ...\\
    \midrule
    \centering{$p_l$} & 
    \centering{/}  &      
    Question: What is the country of citizenship of Warren Buffett? \newline
    $t_l$: United States of America\newline
    Alias: the United States, the United States of America, ...
    \newline
     \\
    \midrule
    \centering{$p^T_g$} & 
    \centering{/} &      
    Question: Which language was Invisible Man written in? \newline
    $t^T_g$: Sanskrit\newline
    Alias: Samskrta, Samskrtam, Sanskrit language, ... \newline
     \\
    \bottomrule
    
    \end{tabular}
    }
    \caption{SRO\_edit multimodal input examples.}
    \label{tab:sro}
\end{table}

\begin{table}
    \centering\small
    \setlength{\tabcolsep}{4mm}{
    \begin{tabular}{m{10pt}m{0.5in}m{1.5in}}
    \toprule
    \centering
    Input & \makecell[c]{Visual\\ Inputs} & Textual Inputs \\
    \midrule
    \centering{Edit\\input} & 
    \includegraphics[width=0.5in]{figures/image1.jpg} &      Question: The official work language of the book in the picture has changed.\newline
    The book in the picture was written in the language of \newline
    $\tilde{o}$: Sanskrit\newline
    \\
    \midrule
    \centering{$p_r$} & 
    \includegraphics[width=0.5in]{figures/image1.jpg} &      Question: The book in the picture was written in the language of \newline
    $t_r$: Sanskrit\newline
    Alias: Samskrta, Samskrtam, Sanskrit language, ... \\
    \midrule
    \centering{$p_c$} & 
    \centering{/} &      
    Question: Invisible Man was written in the language of \newline
    $t_c$: Sanskrit\newline
    Alias: Samskrta, Samskrtam, Sanskrit language, ...\\
    \midrule
    \centering{$p_l$} & 
    \includegraphics[width=0.5in]{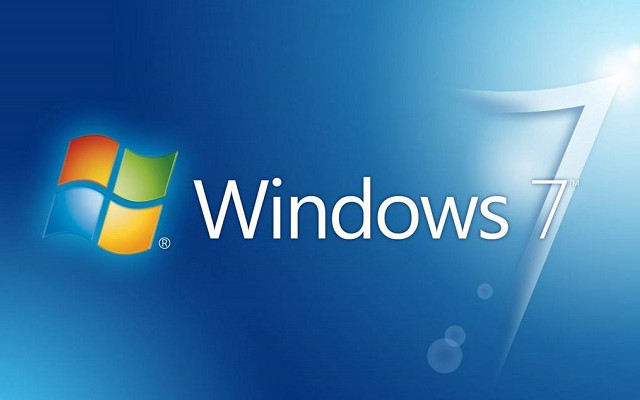} \newline
     &      
    Question: Who is the developer of the operating system in the picture? \newline
    $t_l$: Microsoft\newline
    Alias: MSFT, Microsoft Corp., ...
    \newline
     \\
    \midrule
    \centering{$p^M_g$} & 
    \includegraphics[width=0.5in]{figures/image_gen.jpg} \newline
     &      
    Question: The book in the picture was written in the language of \newline
    $t^M_g$: Sanskrit\newline
    Alias: Samskrta, Samskrtam, Sanskrit language, ...\\
    \midrule
    \centering{$p^T_g$} & 
    \includegraphics[width=0.5in]{figures/image1.jpg} \newline &      
    Question: Which language was the book in the picture written in? \newline
    $t^T_g$: Sanskrit\newline
    Alias: Samskrta, Samskrtam, Sanskrit language, ...\\

    \bottomrule
    
    \end{tabular}
    }
    \caption{IRO\_edit multimodal input examples.}
    \label{tab:iro}
\end{table}

\end{document}